\documentclass[mlmain]{jmlr}

\usepackage{booktabs}

\newcommand{\crProjRatioSixteen}{88\%}

\newcommand{\crFullAffMin}{0.57}
\newcommand{\crFullAffMax}{0.91}

\newcommand{\crProjAffCircMin}{3\%}
\newcommand{\crProjAffCircMax}{42\%}
\newcommand{\crProjGradCircMin}{1\%}
\newcommand{\crProjGradCircMax}{23\%}

\newcommand{\crFwGainMin}{+0.008}
\newcommand{\crFwGainMax}{+0.033}

\newcommand{\crFwShareMed}{3.4\%}
\newcommand{\crFwShareMedRound}{3\%}

\newcommand{\crOccFwShareMin}{2\%}
\newcommand{\crOccFwShareMax}{25\%}
\newcommand{\crOccFwShareMed}{17\%}
\newcommand{\crOccFwPositions}{64{,}000}
\newcommand{\crPosReal}{48\%}
\newcommand{\crPosNull}{49\%}

\newcommand{\crPresWikiSmall}{0.75}

\newcommand{\crPresPileSmall}{0.67}
\newcommand{\crPresWikiLarge}{0.77}
\newcommand{\crDensIccMin}{0.79}
\newcommand{\crDensIccMax}{0.87}

\newcommand{\crNormFourTen}{+0.47}
\newcommand{\crPartFourTen}{+0.59}

\newcommand{\crOccPositions}{3{,}024{,}594}
\newcommand{\crOccQuadWins}{12}

\newcommand{\crOccGradMean}{0.05}
\newcommand{\crGradCeiling}{0.90}

\newcommand{\crPileQuadWinsMin}{12}
\newcommand{\crPileGradMean}{-0.006}
\newcommand{\crPileGradMax}{+0.094}
\newcommand{\crIdPathMin}{0.93}
\newcommand{\crIdPathMax}{1.19}
\newcommand{\crIdPathFused}{0.47}
\newcommand{\crIidMin}{0.45}
\newcommand{\crIidMax}{0.91}

\newcommand{\crRidgeExitLoose}{19\%}

\newcommand{\crHelmMin}{4\%}

\graphicspath{{figures/}}
\hypersetup{pdftitle={The Depth Flow of Token Representations Is Nonlinear and Does Not Descend Its Own Density},
  pdfauthor={Alexandre Quemy}}
\jmlrvolume{}
\firstpageno{1}
\jmlrproceedings{}{}   
\jmlryear{2026}
\jmlrworkshop{Symmetry and Geometry in Neural Representations}

\title[The Depth Flow of Token Representations]{The Depth Flow of Token
Representations Is Nonlinear and Does Not Descend Its Own Density}

\author{\Name{Alexandre Quemy} \Email{alexandre@hother.io}\\
 \addr Hother Labs\\
 Centre for Credible AI, Warsaw University of Technology}

\begin{document}
\maketitle

\begin{abstract}
A token's representation is carried through the network layer by
layer. The whole vocabulary carried together forms a flow. We fit this
flow's equation of motion as a discrete Langevin model over
corpus-mean trajectories of Pythia-160M and Pythia-410M, and score the
predicted steps on held-out tokens. 
Linear maps are often used as cheap surrogates for a layer. The flow
they summarize is not linear: a quadratic drift beats the
linear map at every transition of both models, and the
Kramers--Moyal estimator agrees wherever its neighborhoods stay local.
At full width, the step of a mean trajectory is nearly affine,
\crFwShareMedRound{} non-affine, while the step of an individual state
is not, \crOccFwShareMed{}.
 We then characterize the flow further.
First, we show that it does not descend its own log-density,
whatever the diffusion matrix.
The drift instead descends a potential that is not the
density. Second, the rotational component is not negligible, $4$ to
$45\%$ of the explainable drift, and it is the antisymmetric part
of the linear map. Finally, a token keeps its angular rank and its
local-density rank across all thirteen layers, on two corpora and both
models.
\end{abstract}

\section{Introduction}
\label{sec:intro}
\begin{figure}[t]
\centering
\includegraphics[width=0.82\linewidth]{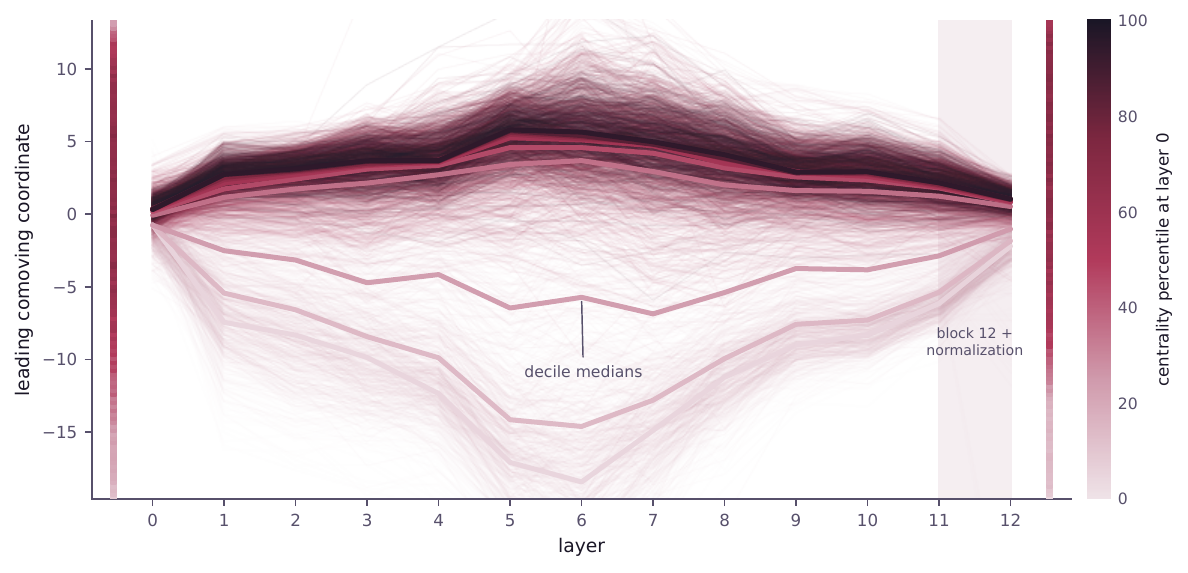}
\caption{Token trajectories, drawn along the leading comoving
coordinate, the axis of largest variation once each layer's cloud is
standardized, so a token's height is its position relative to all tokens. 
Trajectories are colored by the token's centrality percentile at
layer $0$, its rank in cosine to the ensemble mean direction, bold lines are decile medians, and the side strips give the
median color at each height at entry and exit.}
\label{fig:river}
\end{figure}

A token's representation is transformed layer by layer via the residual stream, from its embedding row to the state 
vector used by the model to predict the next token.
Across depth, each token traces a trajectory, and the whole vocabulary together forms a flow as depicted in Figure~\ref{fig:river}.

Writing $s^{\ell}_{t}$ for the residual-stream state at corpus
position $t$ after $\ell$ blocks, and $\mathcal{P}(v)$ for the set of
positions whose next token is $v$, the trajectory of a vocabulary token
$v$ is
\[
  h^{\ell}_{v} \;=\; \frac{1}{|\mathcal{P}(v)|}
  \sum_{t \in \mathcal{P}(v)} s^{\ell}_{t},
  \qquad \ell = 0, \dots, L ,
\]
and the flow is the family of these trajectories over the vocabulary.
A trajectory begins at the mean state of the positions that precede
$v$, not at $v$'s own embedding row. We call it the next-token-conditioned mean trajectory of $v$, and the law we fit is a law for these means.

Concretely, take $v = $ ``dog'' and a corpus that contains ``he walked the dog'' 
and ``she fed her dog''. Then $\mathcal{P}(\text{dog})$ holds the positions of 
``the'' and ``her'', the two moments the model is about to produce ``dog''. 
Each of those positions carries one vector through thirteen states, 
and averaging them state by state gives the 
path $h^{0}{_\text{dog}}, \dots, h^{12}{_\text{dog}}$: at layer $0$ the 
mean embedding of ``the'' and ``her'', at layer $12$ the mean vector from which 
the model says ``dog''. Over the full corpus the average runs over every such 
position.

We are interested in this particular quantity because it is the one vector that the network 
actually carries end to end, from the embedding to the final state. The average 
over the corpus is only pragmatic: as the trajectories are context-dependent, they would be most likely unique, 
such that we would not be able to study them statistically. 

In this paper, we study this flow of token representations across depth, and in particular 
its equation of motion. In other words, how does a model transform a token representation from one layer to the next.
We model the flow as a Langevin equation, with a drift and a diffusion term, and we ask whether the drift is linear or nonlinear, and whether it is a gradient flow or not.
We use three estimators: with Kramers-–Moyal (KM) expansion~\citep{friedrich2011approaching}, a linear map, 
and a quadratic map. The linear map is known to transfer across models and
scales~\citep{yomdin2024jump}.

Our contributions are:
\begin{itemize}
  \item The drift is nonlinear. In a reduced state of $16$ components a quadratic drift beats the linear map at every layer of Pythia-160M and Pythia-410M. At full width, the step of a mean trajectory is nearly affine, \crFwShareMedRound{} non-affine, while the step of an individual state is not, \crOccFwShareMed{}. 
  \item The drift is not the gradient of the log-density. We measure the alignment between the drift and the estimated density gradient and find nothing. 
  We show that the motion has a gradient component, but the diffusion and the rotational part are also load-bearing. 
  \item Among four observable quantities of a token in a cloud of tokens (cosine to the mean direction, vector norm, participation ratio and local density),
  the cosine to the mean direction and the local density are strongly preserved, on two corpora and both models.
\end{itemize}




\section{Related work}

A flow of token representations means three different things in the
literature. The
analytical line treats depth as continuous time and derives the
dynamics rather than measuring them: residual networks as discretized
flows~\citep{chen2018neural}, tokens as interacting particles that
cluster with depth~\citep{geshkovski2025mathematical}, attention as
diffusion~\citep{sander2022sinkformers}. The empirical line follows
residual-stream states across layers for individual prompts and
measures their
geometry~\citep{fernando2025dynamics,fernando2026spectral,disipio2025curved}.
\citet{carson2025statistical} fit a drift-diffusion system, but along
generation time rather than depth. Ours is a fourth object: one mean
trajectory per vocabulary token, over the corpus, along depth, with
the equation fitted from the trajectories and scored on held-out
tokens.

The closest prior work to ours is \citet{sarfati2024lines}. They fit a
stochastic equation to layer-wise trajectories and simulate it. Their
drift is linear by construction, a rotation and stretch, their noise
is isotropic and state-independent, and the gradient reading is never
tested. We test all three: the drift is nonlinear
(Section~\ref{sec:nonlinear}), the noise is correlated and moves with
the state (Section~\ref{sec:nonlinear}), and the gradient reading
fails (Section~\ref{sec:gradient}). The object also differs: they use a trajectory
 per prompt, while we use the mean trajectory per vocabulary
token.

\citet{yomdin2024jump} project hidden states from one layer to another
with a linear map, accurately and across model scales. Their map is
therefore the baseline any law of motion on depth must be compared to, which we do in
Section~\ref{sec:nonlinear}.

Our depth profile is consistent with published geometry. Predictability
rising through depth is the iterative-inference
literature~\citep{jastrzebski2018residual,nostalgebraist2020logit,geva2022promotion,belrose2023tuned}.
Early layers hold information in
superposition~\citep{lombardo2026phases}, and depth proceeds in
stages~\citep{lad2024stages}. The embedding table's geometry is
organized by
frequency~\citep{mu2018allbut,gao2019degeneration,ethayarajh2019contextual,puccetti2022outlier}.

The instruments that we use in this paper are not new either.
The Kramers--Moyal expansion is the standard estimator of drift and
diffusion from sampled paths~\citep{friedrich2011approaching}, its
finite-sampling biases are known~\citep{lade2009finite}, and the
Langevin--Fokker--Planck correspondence is
classical~\citep{risken1989fokker}. The literature already uses the same
equation at least twice, 1) as the continuous limit of stochastic gradient
descent~\citep{mandt2017stochastic} and 2) as the backbone of score-based
generative models~\citep{song2021scorebased}, where the drift is the
score by construction.

\section{The flow and how we measure it}
\label{sec:flow}

This section describes the flow, the estimators for the model and the evaluation protocol. The code, the results and the figures of this paper are at \url{https://github.com/hotherio/paper-neurreps-depth-flow}.

\textbf{The trajectories.} We use $25{,}000$ WikiText-103 articles
that are longer than $400$ characters, truncated at $128$ tokens for practical reasons.
We streamed them through Pythia-160M~\citep{biderman2023pythia} to collect the thirteen hidden states, 
and accumulate the means $h^{\ell}_{v}$ defined in  Section~\ref{sec:intro} over the corpus. We then filter out 
tokens with fewer than five occurrences. 
The result is a set of $25{,}268$ trajectories of thirteen points in $\mathbb{R}^{768}$.


\textbf{The reduced state.} Because the estimator we will fit uses nearest neighbors, we need to reduce the state dimension. 
Another reason is that increments in different layers are expressed in different coordinates, and we want to compare them.
We fit a principal component basis on a pooled sample of $8{,}000$ tokens across all thirteen layers. 
To account for the fact that the norm of the residual stream grows with
depth, we standardize each layer's cloud to the same size when fitting
the basis, dividing each layer's centered states by that layer's radius,
and then project the raw states through the single fixed basis. We tried 8, 16, 32, and 64 components, 
and report the results at 16, which is the working width of \citet{yomdin2024jump}.
At 16, the PCA retains $82.7\%$ of the pooled variance,
which is not uniform across layers: $12\%$ of layer $0$'s own variance,
$79\%$ mid-network, $52\%$ at the output. 

\textbf{The model.} We model the trajectory of every token as a discrete-time Langevin equation with layer-specific coefficients. 
Writing $z^\ell_v \in \mathbb{R}^{16}$ for token $v$'s state at depth $\ell$ 
(the 768-d mean state projected onto the shared 16-component basis), the model is

\[ z^{\ell+1}_v = z^\ell_v + a_\ell(z^\ell_v) + \sqrt{D_\ell(z^\ell_v)}\,\xi^\ell_v, \qquad \xi^\ell_v \sim \mathcal{N}(0, I), \]
with $\xi^\ell_v$ drawn independently for every token and every
transition.

Note that the push $a_\ell$ and the noise scale $D_\ell$ do not depend
on the token $v$, since our hypothesis is that the flow is a property of
the model: they are fixed functions of position, and a token enters only
through its current position $z^\ell_v$ and its own draw $\xi^\ell_v$.
$D_\ell$ is the covariance of what the drift leaves unexplained.

\textbf{The estimators.} Let
$\Delta_v = h^{\ell+1}_v - h^\ell_v$ be a token's step. We consider three models to
estimate the push $a_\ell$ and the noise scale $D_\ell$ from the training tokens:
\begin{itemize}
\item \emph{The local reference.} To predict the step at a state $x$,
  the Kramers--Moyal expansion~\citep{friedrich2011approaching} finds
  the $k = 64$ nearest training states and uses their steps directly,
  their mean as the push and their covariance as the noise,
  \[
    a(x) \;=\; \frac{1}{k}\!\!\sum_{u \in \mathcal{N}_k(x)}\!\! \Delta_u ,
    \qquad
    D(x) \;=\; \frac{1}{k}\!\!\sum_{u \in \mathcal{N}_k(x)}\!\!
    \big(\Delta_u - a(x)\big)\big(\Delta_u - a(x)\big)^{\!\top} .
  \]
 
\item \emph{The affine map.} One matrix and one offset per transition,
  $h^{\ell+1} = A_\ell h^\ell + b_\ell$, fitted by least squares. This
  is the baseline of \citet{yomdin2024jump}. The noise is the covariance of the residuals.
\item \emph{The quadratic map.} The same least-squares fit, with the
pairwise products of the coordinates added as features. The noise is the covariance of the residuals.
\end{itemize}

We include nulls for control: fixed Gaussians that never look at
the state. Null A fits the mean step of each transition with a
per-coordinate variance, null B shares one step distribution across all
transitions, and null C gives null A a full covariance.

\textbf{The protocol.} All fits use one $80/20$ token split. 
The target is: given a held-out token's state at layer $\ell$, predict its step to layer
$\ell + 1$ as a Gaussian, mean from the drift, covariance from the
noise model. We use two metrics: 1) $R^2$ to measure
the drift alone, as the fraction of step variance the predicted push
explains and 2) mean log-likelihood.

\textbf{The noise floor.} Because we average over occurrences, the trajectories' noise is directly connected to frequency.
To measure this noise we split each token's occurrences into two random halves at the final layer and compare the two half-averages, 
which agree only as well as the full average is reliable. For a token seen five times, the resulting noise is $40\%$ of the typical 
distance between different tokens' positions. It falls to $21\%$ for tokens seen ten to a hundred times, and to $2\%$ above a thousand.
The problem is that the trajectories of rare tokens are noisier by construction which remains one of the limitations of our current protocol. 


\section{The drift is nonlinear}
\label{sec:nonlinear}

Both non-linear estimators beat the linear map. At the
working width of $16$ components, the quadratic map wins all twelve
transitions in held-out variance explained (Table~\ref{tab:quad}), and
the local law wins eleven of twelve ($R^2$ of $0.54$ to $0.96$ against
$0.49$ to $0.91$) and all twelve matched-noise likelihood contests
(Table~\ref{tab:sde} in Appendix~\ref{app:likelihood}). Figure~\ref{fig:quadfield} shows the same
bend seen by both where the affine field stays rigid.

Scored against three state-blind nulls, a
per-transition Gaussian (A), the same pooled across transitions (B),
and the same with full covariance (C), the KM expansion beats A and B on
all twelve transitions by $8.4$ to $29.1$ nats (Table~\ref{tab:sde}).

The diffusion part is important. When both models use the same diagonal noise, the KM expansion beats
the linear map twelve of twelve. When the linear map alone gets a full
covariance, it wins eight of twelve, and even null C, which has no
drift, wins three: the richer noise model wins by itself. When both
models get the full covariance, the KM expansion again wins all
twelve, by $1.1$ to $7.6$ nats, with bootstrap intervals excluding
zero, at three seeds. 

\begin{figure}[t]
\centering
\includegraphics[width=1\linewidth]{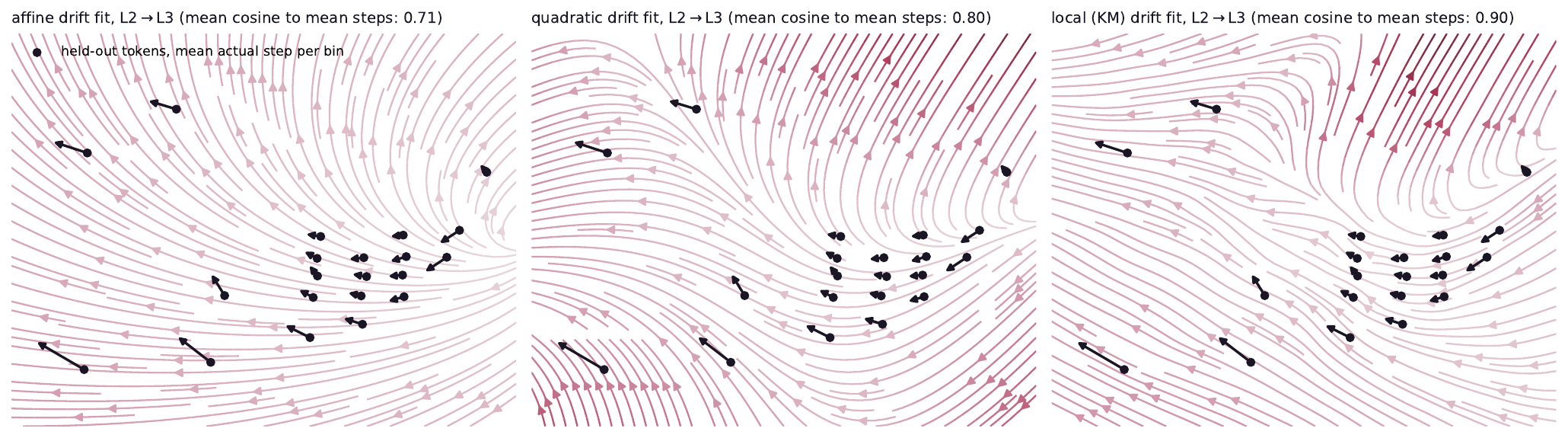}
\caption{The transition from layer $2$ to layer $3$ in the leading
principal plane of layer $2$. In each panel the drift is fitted on these
two coordinates only, by the affine map, the quadratic map and the local
estimator. Black arrows are the mean held-out step in each bin of the
plane.}
\label{fig:quadfield}
\end{figure}

The KM expansion degrades with width: twelve wins at $8$ components,
five at $32$, one at $64$. Its fixed $64$-point neighborhood grows
from $11\%$ to $28\%$ of the cloud radius, and raising $k$ to $256$ or
$1{,}024$ gives zero wins. This is a failure of the estimator, not a
fact about the flow.
The quadratic map wins twelve of twelve at all
four widths, with medians of $0.73$ to $0.85$ against the linear map's
$0.57$ to $0.81$ (Table~\ref{tab:quad}), and all twenty-four
transitions of Pythia-410M at $8$, $16$, and $32$ components. At $64$
on 410M it wins $9$ of $24$: it needs $2{,}145$ coefficients and has
$9{,}549$ trajectories. 160M subsampled to the same budget shows the
same drop, $12$ of $12$ becoming $4$ of $12$ at $64$, unchanged at
$32$. The largest margin is on the final transition, where the
normalization is itself nonlinear. Without it the quadratic still wins
$11$ of $11$, and $23$ of $23$ on Pythia-410M.  We believe that with a larger corpus, 
and therefore more data points, KM would perform the best at every scale. We leave this to future work.

\begin{table}[t]
\centering
\caption{Quadratic drift against the affine map, both parametric and both
gaining capacity with the state width. Wins are transitions where the
quadratic drift explains more held-out increment variance ($12$
transitions on Pythia-160M, $24$ on Pythia-410M). $R^2$ values are
medians over transitions. The last block subsamples the Pythia-160M
training set to Pythia-410M's budget, reproducing the $64$-component
drop and identifying it as an estimation limit.}
\label{tab:quad}
\small
\begin{tabular}{llccccc}
\toprule
Model & Train pts & Width & Quad params & Wins & $R^2$ quad & $R^2$ affine \\
\midrule
Pythia-160M & $20{,}214$ & 8 & 45 & 12/12 & 0.73 & 0.57 \\
Pythia-160M & $20{,}214$ & 16 & 153 & 12/12 & 0.83 & 0.69 \\
Pythia-160M & $20{,}214$ & 32 & 561 & 12/12 & 0.84 & 0.76 \\
Pythia-160M & $20{,}214$ & 64 & 2145 & 12/12 & 0.85 & 0.81 \\
\midrule
Pythia-410M & $9{,}549$ & 8 & 45 & 24/24 & 0.76 & 0.67 \\
Pythia-410M & $9{,}549$ & 16 & 153 & 24/24 & 0.83 & 0.76 \\
Pythia-410M & $9{,}549$ & 32 & 561 & 24/24 & 0.85 & 0.78 \\
Pythia-410M & $9{,}549$ & 64 & 2145 & 9/24 & 0.79 & 0.79 \\
\midrule
Pythia-160M & $9{,}549$ & 32 & 561 & 12/12 & 0.83 & 0.76 \\
Pythia-160M & $9{,}549$ & 64 & 2145 & 4/12 & 0.80 & 0.81 \\
\bottomrule
\end{tabular}
\end{table}

\textbf{What the reduction does.} The reduced state keeps $16$ of
$768$ coordinates, and a step that is affine in all $768$ can still look
curved in $16$ of them. We checked this with a flow that is affine at
full width by construction: a $768$-d affine map fitted per transition,
applied to the real current states, plus the affine residual of a random
other token as noise. In the reduced state it reads almost like the real
flow: the quadratic wins $12$ of $12$ transitions up to $32$ components,
with \crProjRatioSixteen{} of the real margin at $16$ components
(Table~\ref{tab:projection} in Appendix~\ref{app:likelihood}). The same
construction on a Gaussian cloud
with the real mean and covariance gives the opposite result, $0$ of
$12$: through $16$ coordinates an affine flow stays affine when the
cloud is Gaussian, and looks curved only because the real cloud is bent.
The curvature we see in the reduced state is thus mostly the shape of
the cloud, not the map. At full width there is little curvature left to
find: an affine map explains \crFullAffMin{} to \crFullAffMax{} of the
step of the mean trajectories, and letting the map be nonlinear raises
this by only \crFwGainMin{} to \crFwGainMax{}. The average over contexts
removes the context-specific part of each block, and the nonlinearity is
there: on \crOccFwPositions{} individual states the same contest reads a
non-affine share of \crOccFwShareMin{} to \crOccFwShareMax{}, median
\crOccFwShareMed{}, against \crFwShareMed{} for the means. On the Pile the quadratic again wins
\crPileQuadWinsMin{} of $12$ at every width, and \crOccQuadWins{} of
$12$ on \crOccPositions{} un-averaged positions.

We test whether the law can generate whole trajectories. Each held-out
token starts at its real layer-$0$ state, and we apply the fitted law
twelve times in a row, each step adding the predicted push and a draw
from the fitted noise, never touching the real intermediate states.
The tokens it produces end up too close together: their spread is
$0.53$ of the real one mid-network, $0.87$ by the last raw layer, and
a quarter after the final transition. The mid-network figures match the
few-layer rollouts of \citet{sarfati2024lines}. Two artifacts could
explain the collapse, and both are ruled out. Iterated local averaging
could shrink any cloud by itself, but the same pipeline inflates a
stationary control to $2.0$, and the composed linear maps inflate to
$1.7$ rather than shrink.
The likely cause is that the fitted noise mixes persistent token
identity, the structure that Section~\ref{sec:conservation} will show is
preserved, with genuine step-to-step randomness, so the rollout
re-draws at every step what a real token keeps. When each
simulated token keeps one real token's residual path instead, with the
same drift and the same marginal noise size, the spread is
\crIdPathMin{} to \crIdPathMax{} of the real one on the eleven raw
layers, against \crIidMin{} to \crIidMax{} with independent draws, and
only the fused final step still halves it (\crIdPathFused). A law with
a per-token component held fixed along the trajectory is left to future
work.

\begin{figure}[t]
\centering
\includegraphics[width=1.\linewidth]{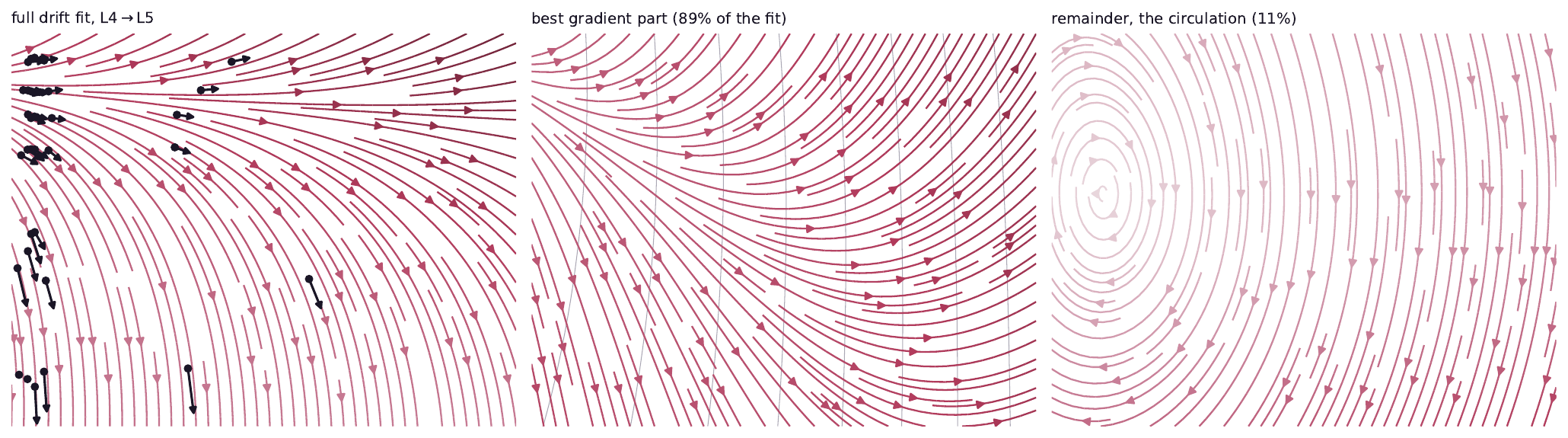}
\caption{One transition split into descent and circulation. Left: the
drift fitted in the leading principal plane of layer $4$, with the mean
actual held-out step per spatial bin in black. Middle: the best
gradient part of that fit, over the contours of its potential. Right:
the remainder, a rotation that no potential can produce. Shares are
held out. In a single plane most of the circulation is invisible, and
the field-level split of Section~\ref{sec:gradient} is the evidence.}
\label{fig:helm}
\end{figure}

\section{The drift is not the gradient of the log-density}
\label{sec:gradient}
We write the drift as a gradient of the log-density:
\[
  a(x) \;=\; D(x)\,\nabla \log p(x) ,
\]
If this reading is true, it would mean that a layer is diffusing particles towards dense regions. 
This is a different view from the energy-descent literature~\citep{zimin2026yuriiformer} where a layer is an optimizer: 
attention descends an interaction energy between the positions, the MLP descends a potential.
Both could be true since the energy-based approach is about what the architecture computes given a context, 
while our approach is concerned with the log-density over the whole vocabulary.

We estimate the density gradient direction at every token from its $k = 48$ 
nearest neighbors  and
measure its mean cosine with the token's actual increment. 
The reading fails at every transition. Against a shuffle null that
pairs each increment with another token's gradient, the mean alignment
gap is $-0.008$, with per-transition gaps between $-0.116$ and
$+0.053$.
The negative results are valid for all state dimensions (8 to 64 components) 
and for both Pythias' parameter counts (Table~\ref{tab:grad} in Appendix~\ref{app:grad}).
To validate the negative result, we create a null and calibrated matched process (Appendix~\ref{app:grad}).
The result does not depend on the diffusion matrix. If $a =
D\,\nabla\log p$ for some positive-definite $D$, then $a \cdot
\nabla\log p > 0$ at every state, whatever $D$ is. On controls with an
isotropic, an anisotropic and a state-dependent $D$, the product is
positive for every token. On the real flow it is positive for
\crPosReal{} of the tokens, and for \crPosNull{} when each increment is
paired with another token's gradient (Appendix~\ref{app:density}). The
null holds on the Pile too (mean gap \crPileGradMean, largest
\crPileGradMax), and on the un-averaged positions (mean gap
\crOccGradMean{} against \crGradCeiling{} on the control).

The negative result for the log-density does not mean the drift is not the gradient of some potential. 
The best cubic potential fitted to the centered drift recovers $55$ to $96\%$ of what the unconstrained
quadratic field explains, leaving $4$ to $45\%$ of rotational part
(Figure~\ref{fig:helm}). We controlled where this rotation comes
from with the two surrogate flows of Table~\ref{tab:projection}. The
flow that keeps the real $768$-d linear map and drops the nonlinearity
rotates as much as the real flow, \crProjAffCircMin{} to
\crProjAffCircMax{}. The flow that keeps only the symmetric part of
that map, which cannot rotate at full width, still reads
\crProjGradCircMin{} to \crProjGradCircMax{} in the reduced state. The
rotation is thus in the linear map of the network, and the reduction
adds at most \crProjGradCircMax{} on top. The rotational part itself is
one minus that ratio, $1 - R^2_{\mathrm{grad}}/R^2_{\mathrm{quad}}$,
with both $R^2$ held out. Its value moves with the ridge of the
potential fit, \crHelmMin{} at the last transition becoming
\crRidgeExitLoose{} at a hundred times the ridge, but the order of the
transitions does not. At linear order, the symmetric share of the
fitted map is $0.53$ to $0.84$ against shuffle nulls of $0.50$ to
$0.54$, largest at the final layer, as in \citet{fernando2026spectral}.

The consequence is that no single landscape summarizes the flow: the
drift mostly descends a potential that is not the log-density, the
remainder circulates, and the correlated noise carries the rest of the
law. This is coherent with what we observe in
Section~\ref{sec:conservation} and in Figure~\ref{fig:river}, where the
rank of the cosine to the mean direction and the rank of the local
density are preserved by the flow while it moves everything.

\section{What the flow preserves}
\label{sec:conservation}

For each layer, we measure four observables: the
cosine to the ensemble mean direction, the vector norm, the
participation ratio $(\sum_i h_i^2)^2 / \sum_i h_i^4$, and the local
density, minus the distance to the tenth nearest token at that layer.

We calculate the intraclass correlation of the token population,
\[
  \mathrm{ICC}(f) \;=\;
  \frac{\operatorname{Var}_v\big[\mathbb{E}_\ell\, \zeta^\ell_v(f)\big]}
  {\operatorname{Var}_v\big[\mathbb{E}_\ell\, \zeta^\ell_v(f)\big]
   \;+\; \mathbb{E}_v\big[\operatorname{Var}_\ell\, \zeta^\ell_v(f)\big]} ,
\]
where $\zeta^\ell_v(f)$ is the per-layer z-score of the observable
$f$ for token $v$, distinct from the reduced state $z^\ell_v$ of
Section~\ref{sec:flow}.

The cosine is preserved with a Spearman rank correlation of $0.76$
between entry and exit, and an ICC of \crPresWikiSmall{} on
WikiText, \crPresPileSmall{} on the Pile and \crPresWikiLarge{} on
Pythia-410M. The local
density is preserved at least as much, with an ICC of \crDensIccMin{}
to \crDensIccMax{} on the same three sets.
On Pythia-160M, the norm is shuffled $-0.17$ while participation ratio is partly reversed at $-0.59$.
The shuffle happens during the layers, as opposed to the participation ratio that is being reversed only at the last layer.
On Pythia-410M both are positively preserved (\crNormFourTen{} and
\crPartFourTen). Therefore, further models and model sizes are needed to
confirm these results.

\section{Conclusion}

We fitted an equation of motion to the depth flow of token
representations: $25{,}268$ corpus-mean trajectories over thirteen
states of Pythia-160M and Pythia-410M, a discrete Langevin model, three estimators,
all scored on held-out tokens.

We showed three things. First, the drift is nonlinear: the quadratic drift
beats the linear map on Pythia-160M and Pythia-410M in a reduced state, and at full width individual states keep a nonlinear part of \crOccFwShareMed{} of their step. Only the mean trajectories are nearly affine at full width. Second, the drift does not
descend its own log-density: the alignment reads nothing at every
width and on both models, against a calibrated instrument, while the
best cubic potential shows the drift mostly descends some other
landscape and $4$ to $45\%$ of it circulates. Finally, the flow preserves a token's angular rank and its
local-density rank across all thirteen layers, and on Pythia-160M
its norm rank is shuffled and its concentration rank is reversed by the
last block.

However, our findings have several limitations. First, we studied only one family of model (Pythia) and at small scale.
Second, the corpus is likely to be too small for the Kramers–Moyal expansion to properly shine.
Third, the law is a regression of the increments: its noise term is
what the drift leaves unexplained, not a diffusion we have established.
Finally, we do not know which part of the block, attention or the MLP,
carries the nonlinear remainder. Therefore,
the next immediate steps are to replicate the study with different families of models, possibly at larger scale.

Another limitation is against the literature. The fitted stochastic differential equation of
\citet{sarfati2024lines} effectively serves as a surrogate of the trajectories,
while ours does not work yet as such. Section~\ref{sec:nonlinear} shows that the fitted noise contains
token identity, and that keeping it fixed along the rollout restores the
spread. Solving this problem would open the door to the main practical
application of such motion law, namely a better surrogate. Another axis around the same question is the study of un-averaged trajectories.
Finally, it remains to explore the identity of the potential that the drift does descend.
We showed the component exist, but we have not identified it yet.

To conclude, token representations do not slide down a landscape into their predictions: they are pushed by a field that keeps their angular order while it moves the whole vocabulary.



\bibliography{references}

\appendix
\section{The gradient test and its calibration}
\label{app:grad}

\textbf{A null result needs a floor.} A near-zero alignment could mean
the flow is not a gradient flow, or it could mean the instrument cannot
see one. The control samples $25{,}268$ points from a stationary
Gaussian with per-coordinate scales log-spaced between $1$ and $4$,
sets $a = D\,\nabla\log p$ with $D = I$, and takes one Euler step per
point at a chosen drift-to-noise ratio, matched to the real flow in
sample size, dimension, and $k$. The estimated gradient direction
agrees with the analytic one at cosine $0.914$. Across drift-to-noise
ratios from $0.9$ to $106$ it reads alignments of $0.61$ to $0.91$
against shuffle nulls within $0.004$ of zero, and in energy terms
places $43$ to $81\%$ of drift energy along the estimated gradient,
against $6$ to $29\%$ on the real flow. A gradient flow would have lit
this instrument up.

\begin{table}[htbp]
\centering
\caption{The gradient reading and its floor. For each state width the
same estimator is run on the real flow and on a matched
Ornstein--Uhlenbeck process for which $a = D\,\nabla\log p$ holds by
construction. Real values are means over the twelve transitions of the
alignment gap over a shuffle null.}
\label{tab:grad}
\small
\begin{tabular}{lcccc}
\toprule
Width & 16 & 32 & 64 & 128 \\
\midrule
Real flow, mean gap        & $-0.008$ & $-0.009$ & $-0.010$ & $-0.004$ \\
Real flow, largest gap     & $+0.053$ & $+0.056$ & $+0.053$ & $+0.050$ \\
Ornstein--Uhlenbeck control & $0.898$ & $0.854$ & $0.819$ & $0.791$ \\
\bottomrule
\end{tabular}
\end{table}

\textbf{And the reading survives the common translation.} Most of each
increment is a shared translation of the whole cloud, $76\%$ of
increment energy in median, and a translation carries no alignment of
its own, so a residual gradient component could hide beneath it:
injecting that translation share into the control drops its reading
from $0.90$ to $0.42$. Removing each transition's mean step restores
the control to $0.90$, and the real flow still reads nothing on the
token-specific residuals, a mean gap of $-0.02$ with a maximum of
$+0.07$.

\section{The reading for any diffusion matrix}
\label{app:density}
Table~\ref{tab:density} gives the sign test of
Section~\ref{sec:gradient} at four widths. The controls are stationary
Gaussian clouds with per-coordinate scales log-spaced between $1$ and
$4$, one Euler step at drift-to-noise ratios $2$ and $5$, with the drift
$D\,\nabla\log p$ for three choices of $D$: the identity; a constant
matrix with eigenvalues log-spaced from $1$ to $16$ in a basis rotated
against the cloud's axes; and $D(x) = (1 + |x|^2/c)\,D_0$ with the
divergence term added to the drift so that the Gaussian stays
stationary. The drift of a held-out token is the local mean step of its
$64$ nearest training tokens and the score direction the unit mean shift
of its $48$ nearest tokens.

\begin{table}[t]
\centering
\caption{The sign test at four widths: the fraction of held-out tokens whose drift has a positive inner product
with the estimated score. Controls is the smallest value over the six controls of Appendix~\ref{app:density}, real and
shuffled are means over transitions, then the per-transition range, the same with the common translation removed, and
the cosine gap once the score is multiplied by the fitted residual covariance.}
\label{tab:density}
\small
\setlength{\tabcolsep}{4pt}
\begin{tabular}{lcccccc}
\toprule
width & controls & real & shuffled & per transition & translation removed & cos.\ gap with $\hat D$ \\
\midrule
16 & 1.00 & 0.479 & 0.490 & 0.37--0.57 & 0.32--0.59 & -0.012 / +0.055 \\
32 & 1.00 & 0.483 & 0.494 & 0.41--0.59 & 0.30--0.61 & -0.013 / +0.041 \\
64 & 1.00 & 0.483 & 0.495 & 0.40--0.59 & 0.28--0.60 & -0.010 / +0.039 \\
128 & 1.00 & 0.489 & 0.500 & 0.39--0.59 & 0.29--0.60 & -0.010 / +0.041 \\
\bottomrule
\end{tabular}
\end{table}

\clearpage
\section{The likelihood contests and the projection control}
\label{app:likelihood}
Table~\ref{tab:sde} is the per-transition likelihood table of
Section~\ref{sec:nonlinear}, Table~\ref{tab:projection} the projection
control of the same section, and Table~\ref{tab:fullwidth} its full-width
contest on the mean trajectories and on individual states.

\begin{table}[t]
\centering
\caption{Quadratic against affine map on the real flow and on four surrogates whose full-width step is known.
Wins are transitions where the quadratic map explains more held-out increment variance, margins are medians over
transitions, and the last column is the range of the circulation share at $16$ components. Cross-fitted: every
training token predicted by a map fitted on the other half.}
\label{tab:projection}
\small
\setlength{\tabcolsep}{4pt}
\begin{tabular}{lcccccc c}
\toprule
& \multicolumn{2}{c}{16 components} & \multicolumn{2}{c}{32} & \multicolumn{2}{c}{64} & circulation (16) \\
flow & wins & margin & wins & margin & wins & margin & \\
\midrule
real flow & 12/12 & +0.083 & 12/12 & +0.068 & 12/12 & +0.042 & 0.04--0.45 \\
affine at full width & 12/12 & +0.074 & 12/12 & +0.061 & 10/12 & +0.018 & 0.03--0.42 \\
affine, cross-fitted & 12/12 & +0.074 & 12/12 & +0.058 & 9/12 & +0.015 & -- \\
affine, Gaussian cloud & 0/12 & -0.002 & 0/12 & -0.006 & 0/12 & -0.021 & -- \\
gradient at full width & 12/12 & +0.014 & 12/12 & +0.009 & 6/12 & +0.001 & 0.01--0.23 \\
\bottomrule
\end{tabular}
\end{table}

\begin{table}[t]
\centering
\caption{The non-affine part of the step per state width, on the mean trajectories and on $64{,}000$ individual
states. Affine $R^2$ is held out; gains are what the network adds to the affine map on the null and on the real flow;
recovery is the fraction of a planted $10\%$ non-affine term the network finds; the last column is the real gain over
the null read through the recovery. Ranges and medians over the twelve transitions of Pythia-160M.}
\label{tab:fullwidth}
\footnotesize
\setlength{\tabcolsep}{4pt}
\begin{tabular}{lcccccc}
\toprule
width & affine $R^2$ & gain on null & gain on real & real $-$ null & recovery & non-affine \\
\midrule
\multicolumn{7}{l}{mean trajectories} \\
raw $768$ & 0.58--0.91 & -0.002 to -0.000 & +0.008 to +0.033 & +0.019 & 0.55 & 3.4\% \\
256 & 0.61--0.93 & +0.004 to +0.033 & +0.020 to +0.060 & +0.024 & 0.51 & 4.5\% \\
64 & 0.58--0.93 & +0.022 to +0.124 & +0.030 to +0.150 & +0.023 & 0.60 & 3.8\% \\
16 & 0.49--0.91 & +0.036 to +0.304 & +0.040 to +0.327 & +0.020 & 0.67 & 2.1\% \\
\midrule
\multicolumn{7}{l}{individual states} \\
raw $768$ & 0.45--0.92 & -0.001 to -0.000 & +0.017 to +0.151 & +0.085 & 0.62 & 16.7\% \\
64 & 0.31--0.88 & +0.040 to +0.161 & +0.061 to +0.453 & +0.089 & 0.47 & 19.6\% \\
16 & 0.20--0.86 & +0.050 to +0.343 & +0.062 to +0.588 & +0.051 & 0.24 & 19.9\% \\
\bottomrule
\end{tabular}
\end{table}

\begin{table}[htbp]
\centering
\caption{Held-out log-likelihood per increment on the twelve transitions
of Pythia-160M, in nats, higher is better, best per row in bold. KM is
the fitted local law with per-coordinate diffusion, KM-full its variant
with a local full covariance ($k = 64$, off-diagonals shrunk by $0.1$).
Null A is the per-transition Gaussian, null B the same pooled across
transitions, null C the full-covariance per-transition Gaussian. Affine
is the least-squares map with per-coordinate or
full residual covariance. The $R^2$ columns give held-out increment
variance explained by each drift.}
\label{tab:sde}
\scriptsize
\setlength{\tabcolsep}{3.2pt}
\begin{tabular}{lccccccccc}
\toprule
Transition & KM & KM-full & Null A & Null B & Null C & Affine &
Affine-full & $R^2$ KM & $R^2$ aff. \\
\midrule
L0$\to$1 & 5.61 & \textbf{8.59} & -8.93 & -44.55 & -2.04 & -0.97 & 2.38 & 0.83 & 0.73 \\
L1$\to$2 & 14.94 & \textbf{16.82} & 4.77 & -43.95 & 10.57 & 11.86 & 14.72 & 0.55 & 0.49 \\
L2$\to$3 & 6.96 & \textbf{11.29} & -4.83 & -44.04 & 3.55 & 4.62 & 8.53 & 0.72 & 0.66 \\
L3$\to$4 & 8.00 & \textbf{9.91} & -5.66 & -45.16 & 2.84 & 4.46 & 7.73 & 0.78 & 0.58 \\
L4$\to$5 & 11.90 & \textbf{14.04} & -1.74 & -43.97 & 9.01 & 9.48 & 12.83 & 0.84 & 0.81 \\
L5$\to$6 & 5.50 & \textbf{12.49} & -2.91 & -43.97 & 7.23 & 2.73 & 10.81 & 0.54 & 0.50 \\
L6$\to$7 & 4.46 & \textbf{11.93} & -4.74 & -44.00 & 5.98 & 2.25 & 10.28 & 0.65 & 0.59 \\
L7$\to$8 & -0.38 & \textbf{6.80} & -12.22 & -44.63 & 0.54 & -1.80 & 5.57 & 0.80 & 0.78 \\
L8$\to$9 & -4.97 & \textbf{2.75} & -17.06 & -46.92 & -4.73 & -5.51 & 1.64 & 0.77 & 0.79 \\
L9$\to$10 & -5.95 & \textbf{0.31} & -16.58 & -45.02 & -8.03 & -8.06 & -1.96 & 0.76 & 0.72 \\
L10$\to$11 & -6.89 & \textbf{-0.97} & -25.88 & -48.63 & -10.80 & -9.01 & -3.49 & 0.92 & 0.91 \\
L11$\to$12 & -14.06 & \textbf{-12.30} & -43.15 & -117.53 & -28.82 & -32.16 & -19.91 & 0.96 & 0.66 \\
\bottomrule
\end{tabular}
\end{table}

\end{document}